\documentclass[conference]{IEEEtran}
\IEEEoverridecommandlockouts    

\usepackage{booktabs}
\usepackage{cite}
\usepackage{amsmath,amssymb,amsfonts}
\usepackage{graphicx}
\usepackage[caption=false]{subfig}
\usepackage{xspace}
\usepackage{algpseudocode}
\usepackage{textcomp}
 \usepackage{array}
\usepackage{hyperref}
\usepackage{setspace}  
\usepackage{tabularx}
\usepackage{adjustbox}
\usepackage{comment}
\usepackage[table]{xcolor} 
\usepackage{makecell}

 \newcounter{algorithm}

\usepackage[normalem]{ulem}
\usepackage{xcolor}

\newcommand{\dataset}{ObjectVisA-120\xspace}

\algnewcommand{\LeftComment}[1]{\Statex \(\triangleright\) #1}

\def\BibTeX{{\rm B\kern-.05em{\sc i\kern-.025em b}\kern-.08em
    T\kern-.1667em\lower.7ex\hbox{E}\kern-.125em}}
    
\begin{document}


\title{\LARGE \bf
ObjectVisA-120: Object-based Visual Attention Prediction in Interactive Street-crossing Environments}

\author{Igor Vozniak$^{1}$, Philipp Müller$^{1,2}$, Nils Lipp$^{1}$, Janis Sprenger$^{1}$, Konstantin Poddubnyy$^{1}$, Davit Hovhannisyan$^{1}$, \\ Christian Müller$^{1}$, Andreas Bulling$^{3}$, Philipp Slusallek$^{1}$
\thanks{$^{1}$German Research Center for Artificial Intelligence (DFKI) GmbH, Campus D32, 66123 Saarbruecken, Germany}%
\thanks{$^{2}$Max Planck Institute for Intelligent Systems, 70569 Stuttgart, Germany}%
\thanks{$^{3}$Institute for Visualization and Interactive Systems (VIS) at
Stuttgart University, 70569 Stuttgart, Germany}%
\thanks{\textbf{Accepted for publication at the IEEE Intelligent Vehicles Symposium
(IV), 2026}}
}

\maketitle
\thispagestyle{empty}
\pagestyle{empty}

\setcounter{footnote}{3}

\begin{abstract}
The object-based nature of human visual attention is well-known in cognitive science, but has only played a minor role in computational visual attention models so far. This is mainly due to a lack of suitable datasets and evaluation metrics for object-based attention. To address these limitations, we present \dataset~\footnote{https://www.kaggle.com/datasets/igorvozniak/objectvisa-120} -- a novel 120-participant dataset of spatial street-crossing navigation in virtual reality specifically geared to object-based attention evaluations. The uniqueness of the presented dataset lies in the ethical and safety affiliated challenges that make collecting comparable data in real-world environments highly difficult. \dataset~ not only features accurate gaze data and a complete state-space representation of objects in the virtual environment, but it also offers variable scenario complexities and rich annotations, including panoptic segmentation, depth information, and vehicle keypoints. We further propose object-based similarity (oSIM) as a novel metric to evaluate the performance of object-based visual attention models, a previously unexplored performance characteristic. 
Our evaluations show that explicitly optimising for object-based attention not only improves oSIM performance but also leads to an improved model performance on common metrics. In addition, we present SUMGraph, a Mamba U-Net-based model, which explicitly encodes critical scene objects (vehicles) in a graph representation, leading to further performance improvements over several state-of-the-art visual attention prediction methods. The dataset, code and models will be publicly released.
\end{abstract}


\section{Introduction} \label{sec:intro}
Visual attention is a fundamental process that allows humans to focus their limited processing resources on the most relevant stimuli in the environment~\cite{chen2012object} and has been extensively studied both in cognitive science~\cite{sood2023improving,de2005semantic, roth2023objects,chun2005visual,roelfsema2006cortical, schneider1995vam} and in computer vision~\cite{Vozniak_2023_WACV, jiang2015salicon, lou2022transalnet, hosseini2025sum, droste2020unified, wang2017deep}.
A major insight from cognitive science is that attention is not only driven by spatial but also object-based factors~\cite{chen2012object,de2005semantic, roth2023objects}.
Object-based attention is guided by the perceptual grouping of features into coherent objects~\cite{vecera1994grouped, egly1994shifting} and allows 
for selective enhancement of entire objects rather than just isolated spatial locations, facilitating more efficient visual search, recognition, and interaction with the environment.
Consequently, cognitive models of visual attention have long incorporated explicit notions of objects~\cite{schneider1995vam,salvucci2001integrated}.
\begin{figure*}[t]
  \centering
   \includegraphics[width=1.\linewidth]{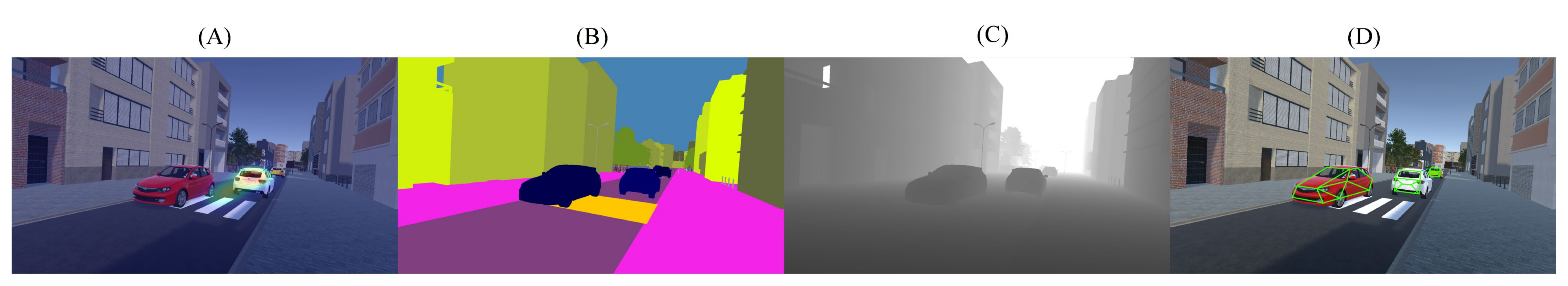}

\caption{\dataset~dataset: (A) displays a FoV image with an overlaid visual attention map during street-crossing navigation task; (B) presents the affiliated panoptic segmentation, following the CityScapes~\cite{Cordts_2016_CVPR} labeling policy; (C) shows the corresponding depth map; and (D)  the keypoints and edges annotations aligned with OpenPifPaf~\cite{kreiss2021openpifpaf} labeling policy.
   }
   \label{fig:dataset}
\end{figure*}
In contrast, research on attention prediction in computer vision largely ignores the object-based nature of human attention allocation.
While recent years have seen tremendous progress in saliency prediction~\cite{jiang2015salicon, lou2022transalnet, hosseini2025sum, droste2020unified,wang2017deep}, these approaches are limited to predicting attention spatially.
Apart from disregarding insights from cognitive science, this focus on spatial attention is also at odds with important requirements of application scenarios.
For instance, when pedestrians cross a busy street, it is crucial to predict whether they are likely to see an important object, such as an approaching vehicle, while their spatial attention distribution is less important. A key reason for this is the lack of suitable metrics to evaluate object-based attention prediction methods.
Existing metrics such as Normalised Scanpath Similarity (NSS), Kullback-Leibler Divergence (KLD), or Similarity (SIM)  only spatially compare predicted and ground truth gaze data without any reference to the objects contained in the scene~\cite{bylinskii2018different}. A major challenge for establishing a novel, object-based evaluation metric is the need for highly accurate object segmentations, which are usually not available in common human attention datasets~\cite{jiang2015salicon, judd2009learning, borji2015cat2000, jiang2022does, jiang2023ueyes, xu2014predicting}. We address these limitations with three major contributions:

\begin{itemize}
    \item \textbf{First}, we present \dataset~(cf. Figure~\ref{fig:dataset}) -- a novel 120-participant human attention dataset recorded using synthetic street-crossing scenarios in virtual reality (VR), therefore allowing access to highly accurate ground truth segmentations of object instances. 
    \item \textbf{Second}, we introduce Object-based Similarity (oSIM), a metric designed to measure the accuracy of attention predictions with respect to scene objects. We show that incorporating oSIM into the training objective consistently improves model performance. 
    \item \textbf{Third}, we introduce SUMGraph, to the best of our knowledge, the first attention prediction model for interactive environments that makes use of explicit object information.
    Evaluations on our novel dataset, demonstrate that SUMGraph outperforms on par with state-of-the-art methods and its ablated versions. 
\end{itemize}

\section{Related Works}\label{sec:related-works}
Human visual attention prediction is studied in a wide variety of scenarios, including scanpath prediction~\cite{chen2021predicting,yang2020predicting}, gaze anticipation~\cite{hu21fixationnet,muller2020anticipating}, or gaze following~\cite{recasens2015they}. Most studies on human visual attention prediction, however, focus on generating context-free saliency maps on still images, using ground-truth maps averaged across multiple observers~\cite{itti2002model, kummerer2016deepgaze, cornia2018predicting, droste2020unified}. Salient features are often extracted by CNN backbone models~\cite{krizhevsky2017imagenet, simonyan2014very, he2016deep,kummerer2017understanding}. The usage of recurrent models like Long-Short Term Memory (LSTM)-based architectures~\cite{cornia2018predicting, liu2018deep}, demonstrated effectiveness in processing both local and long-range visual information. Recently, Transformer-based models~\cite{lou2022transalnet, hosseini2024brand, han2022survey, djilali2024learning} have demonstrated substantial advancements in saliency prediction, achieving state-of-the-art performance by effectively capturing spatial long-range dependencies. SUM \cite{hosseini2025sum} integrates the efficient long-range dependency modeling of Mamba with a U-Net-like architecture, and proposes a Conditional Visual State Space (C-VSS) block which enables token-driven adaptation across various domains and data types (natural scenes, webpages, commercial imagery).
SUM achieved state-of-the-art performance on several saliency prediction datasets, including SALICON \cite{jiang2015salicon}, CAT2000 \cite{borji2015cat2000} and MIT1003 \cite{judd2009learning}. ContextSalNet~\cite{Vozniak_2023_WACV} presents an approach to predict attention distributions in an interactive traffic environment conditioned on task- and personal context information.
While deep neural networks can learn object-sensitive features implicitly~\cite{zhou2014object}, to the best of our knowledge, no previous works on visual attention prediction in interactive environments combined an explicit object representation with deep image features.
Furthermore, due to the lack of an adequate metric and loss function, no previous works were directly optimised for object-based attention prediction.

\textbf{Metrics for evaluating attention prediction.} 
Depending on the precise task formulation, a wide variety of metrics to evaluate predictions of human visual attention are used.
Methods predicting scanpaths (ordered fixation sequences), are often evaluated with sequence comparison metrics such as Dynamic Time Warping or Levenshtein similarity~\cite{fahimi2021metrics}.
More relevant to our work are metrics from the field of saliency prediction, as they compare a predicted attention distribution with ground truth gaze~\cite{bylinskii2018different, kuemmerer2018salmetrics}. 
These metrics can be broadly separated into two categories.
Distribution-based metrics such as Similarity (SIM), Pearson's Correlation Coefficient (CC), and  Kullback-Leibler Divergence (KLD) compare predicted with ground truth attention distributions.
Location-based metrics such as Area Under the Curve (AUC) and Normalized Scanpath Saliency (NSS) compare predicted attention distributions to ground truth fixation locations.
A common feature of all these metrics is that they operate on an abstract image plane without any notion of the attended objects or their semantics.w

\textbf{Datasets.} VR datasets of human gaze behaviour have the advantage of allowing for a wider, more natural field of view compared to what is achievable with screen-based recordings.
In addition, they offer the advantage that gaze behaviour is integrated with natural body movements.
In case the presented scenes are synthetic, it is possible to access highly accurate panoptic segmentation information, which is crucial for evaluating the quality of object-based attention predictions.
In Table~\ref{tab:vr_datasets_compact} we provide an overview of the most popular VR-based datasets focused on saliency prediction in real and synthetic environments.
The free-viewing~\cite{hu2021ehtask, hu21fixationnet, bernal2023d, xu2024panonut360, david2023salient360, xu2018gaze, wang2022salientvr} and search~\cite{hu2021ehtask, hu21fixationnet} tasks are well-supported by existing datasets. The recently introduced MoGaze datasets explored an interaction task~\cite{kratzer2020mogaze}. HOT3D~\cite{banerjee2024hot3d} and AEA~\cite{lv2024aria} are datasets focusing on routine eye-gaze attention with hand movements and navigation/interaction between multiple participants, respectively. 
ContextSalNet~\cite{Vozniak_2023_WACV} explored pedestrian saliency prediction in street-crossing navigation tasks. However, with only 11 participants, this dataset is significantly smaller compared to \dataset~with 120 participants. Moreover, it suffers from limited visual realism and insufficient data annotation support. While offering highly accurate panoptic segmentation annotations would be feasible in the case of synthetic datasets, none of the previous datasets provides this information, making it impossible to effectively study object-based attention on these datasets. Importantly, the street-crossing navigation task involves significant safety and ethical challenges, making it impractical to conduct such studies in real-world environments due to the setup’s complexity and potential pedestrian bias.

\section{\dataset Dataset}
\label{sec:dataset}

\begin{table*}[t]
\centering
\renewcommand{\arraystretch}{1.2}

\caption{The comparison of saliency and gaze prediction datasets (VR).
In the column Task, we use a separate notation to represent the task, where F stands for the free viewing task, S corresponds to the search task, N stands for navigation task, SC denotes street-crossing constraints, and I stands for the interaction task. The (*) symbol indicates partial or limited annotation.}
\label{tab:vr_datasets_compact}

\fontsize{5pt}{6pt}\selectfont 

\begin{adjustbox}{width=\textwidth}
\begin{tabular}{%
p{1.8cm}
>{\centering\arraybackslash}p{0.8cm}
>{\centering\arraybackslash}p{0.3cm}
>{\centering\arraybackslash}p{0.2cm}
>{\centering\arraybackslash}p{0.2cm}
>{\centering\arraybackslash}p{0.2cm}
>{\centering\arraybackslash}p{0.2cm}
>{\centering\arraybackslash}p{0.2cm}
>{\centering\arraybackslash}p{0.2cm}
>{\centering\arraybackslash}p{0.3cm}
>{\centering\arraybackslash}p{2.2cm}
}
\toprule
Dataset  & Real/Synth. & Task & \rotatebox{10}{Depth} & \rotatebox{10}{Segment.} & \rotatebox{10}{Instance} & \rotatebox{10}{Panoptic} & \rotatebox{10}{Skeleton} & \rotatebox{10}{Sound} & Participants & Size \\
\midrule

SalientVR~\cite{wang2022salientvr} & Mixed & F &  &  &  &  &  &  & 45 & 103 videos \\
SaliencyInVR~\cite{sitzmann2018saliency} & Real & F &  &  &  &  & \checkmark* &  & 169 & 22 frames \\
VR-EyeTracking~\cite{xu2018gaze} & Real & F &  &  &  &  &  &  & 45 & 208 videos \\
Salient360!~\cite{david2023salient360} & Real & F &  &  &  &  &  &  & 48 & 60 frames \\
DeepIntoVS~\cite{celikcan2020deep} & Synthetic & F &  &  &  &  &  &  & 50 & 6,978 frames \\
Panonut360~\cite{xu2024panonut360} & Mixed & F &  &  &  &  &  &  & 50 & 15 videos \\
D-SAV360~\cite{bernal2023d} & Real & F & \checkmark &  &  &  &  & \checkmark & 87 & 85 videos \\
DGaze~\cite{hu2020dgaze} & Synthetic & F &  &  &  &  &  &  & 43 & 5 scenes \\
FixationNet~\cite{hu21fixationnet} & Synthetic & S &  &  &  &  &  &  & 27 & 162 trials \\
MoGaze~\cite{kratzer2020mogaze} & Real & I+N & \checkmark* & \checkmark* &  &  & \checkmark &  & 7 & 3 hours \\
EHTask~\cite{hu2021ehtask} & Real & F+S &  &  &  &  &  &  & 30 & 15 videos \\
HOT3D~\cite{banerjee2024hot3d} & AR/VR & I &  &  &  &  & \checkmark* &  & 19 & 3.7M frames \\
AEA~\cite{lv2024aria} & Real & I+N &  &  &  &  &  & \checkmark & 4 & $\sim$1Mil.\ frames \\
ContextSalNet~\cite{Vozniak_2023_WACV} & Synthetic/VR & N+SC & \checkmark  & \checkmark &  &  &  &  & 11 & 35K frames \\
\midrule

\rowcolor{gray!20}\textbf{\dataset\ (Ours)} & Synthetic & N+SC &
\checkmark & \checkmark & \checkmark & \checkmark & \checkmark* & \checkmark & 120 & 7,200 videos, 6.14M frames \\
\bottomrule
\end{tabular}
\end{adjustbox}

\end{table*}

\textbf{Study.} 
This work builds upon an immersive street-crossing study with bidirectional traffic flow within the Unity 3D engine~\cite{sprenger2023crosscdr}. It includes scenes with different levels of traffic density and dynamics, presence and absence of crosswalks, as well as different numbers of non-playable characters with different behaviours (risky vs. cautious) added to influence the participants. The diversity of the setup and resulting dataset is ensured by including 120 participants, emphasizing variation across individual characteristics.
HTC Vive Pro Eye head-mounted VR goggles has been used, equipped with a wireless adapter to enable unrestricted movement within the effective tracking area (9 $\times$ 8 meters footprint). This makes it the first VR dataset of its kind, pushing the boundary between virtual and real environments to ensure both realism and uniqueness. In total, eye-gaze fixations were recorded from 120 participants, representing a diverse range of personal attributes: \textbf{nationality} \textit{(even split between German and Japanese)}, \textbf{age} \textit{(Range: 20–50 years, $\mu$: 30.56, $\sigma$: 9.02)}, \textbf{gender} \textit{(60 male, 60 female)}, \textbf{height} \textit{(Range: 151.50 - 190.50 cm, $\mu$: 172.46 cm, $\sigma$: 9.47 cm)}, \textbf{weight} $\textit{(Range: 41.60 - 119.60 kg, $\mu$: 66.18 kg, $\sigma$: 14.57 kg)}$, \textbf{driving experience in years} \textit{(Range: 0 - 32, $\mu$: 9.42, $\sigma$: 8.85)}, and \textbf{VR familiarity} \textit{(81 yes, 39 no)}.
Each participant performed 60 street-crossing trials~\cite{sprenger2023crosscdr}.
Prior to the study and after eye-tracking sensor calibration, each participant had a warm-up phase to get used to the environment, physical and virtual boundaries, and adjust to the correspondence between their physical movements and virtual navigation.  A detailed description of the study design is available in the corresponding paper \cite{sprenger2023crosscdr}. Importantly, none of the contributions presented in this work overlap with those in~\cite{sprenger2023crosscdr}, which describes the general setup and study design, but neither presents a publicly available dataset for attention prediction, nor proposes a framework for object-based attention prediction.

\textbf{ObjectVisA-120 dataset.} 
While data was recorded at 90Hz, for the purpose of this work, we chose to consider every 3rd frame, resulting in $6.142.167$ frames (sampling rate of 30Hz).
In addition to the raw field-of-view RGB images (cf. Figure~\ref{fig:dataset}, A), the dataset comprises corresponding eye-gaze fixation information (2D/3D) and derived saliency maps (cf. Figure~\ref{fig:dataset}, A overlaid) computed individually, panoptic segmentation (covering vehicles, pedestrians, crossings, road and pavement surfaces, buildings, vegetation, signage, and other street furniture as unary objects), depth maps (cf. Figure~\ref{fig:dataset}, B-C), and skeleton labels (cf. Figure~\ref{fig:dataset}, D) for vehicles. Notable, since the entire state space of the study has been saved, it is possible to generate additional modalities like 2D/3D bounding boxes or add another layer of segmentation. To maintain consistency with previous work, the skeleton labelling corresponds to the sparse variation of the ApolloCar3D dataset as used in OpenPifPaf~\cite{kreiss2021openpifpaf}, which contains 24 keypoints. Incorporating pedestrian skeleton representations is beneficial due to their strong relevance for behavioural and intent inference, whereas extending graph representations to static scene elements offers limited added value and incurs unnecessary computational overhead. The panoptic segmentation on \dataset~follows the CityScapes~\cite{Cordts_2016_CVPR} labeling policy. The extensive labeling of \dataset~enables multiple computer vision tasks to be performed, including human-like trajectory generation empowered by visual attention~\cite{vozniak2020infosalgail}, providing a significant value to the community. Note that the annotation policy for occluded objects closely reflects real-world perception, where the occluded parts of objects are excluded from all labels (cf. Figure~\ref{fig:dataset}). Where necessary, state-of-the-art methods may be used to generate on-demand segmentation labels for real-world datasets and applications. Subsequent fine-tuning on the ApolloCar3D dataset can be applied to derive the required skeleton representations of vehicles, people and animals supporting deployment in real-world settings. To generate ground truth attention maps, we follow previous work~\cite{Vozniak_2023_WACV}, where 2D sparse ground truth attention maps were created using the last three fixation points.  According to~\cite{wang2023foveated}, the eccentric angle of the foveal region is approximately 2$^{\circ}$, additionally, the eye-tracking accuracy of VR headset\footnote{HTC Vive Pro Eye: https://developer.vive.com/resources/hardware-guides/vive-pro-eye-specs-user-guide/} ranges from 0.5$^{\circ}$-1.1$^{\circ}$. Thus, the ground-truth attention maps were generated using Gaussians with a standard deviation of $3\times dva$ (degree of visual angle).

\section{Object-based Similarity (oSIM) Metric}
\label{sec:osim}

\begin{figure}[h]
  \centering
   \includegraphics[width=1.0\linewidth]{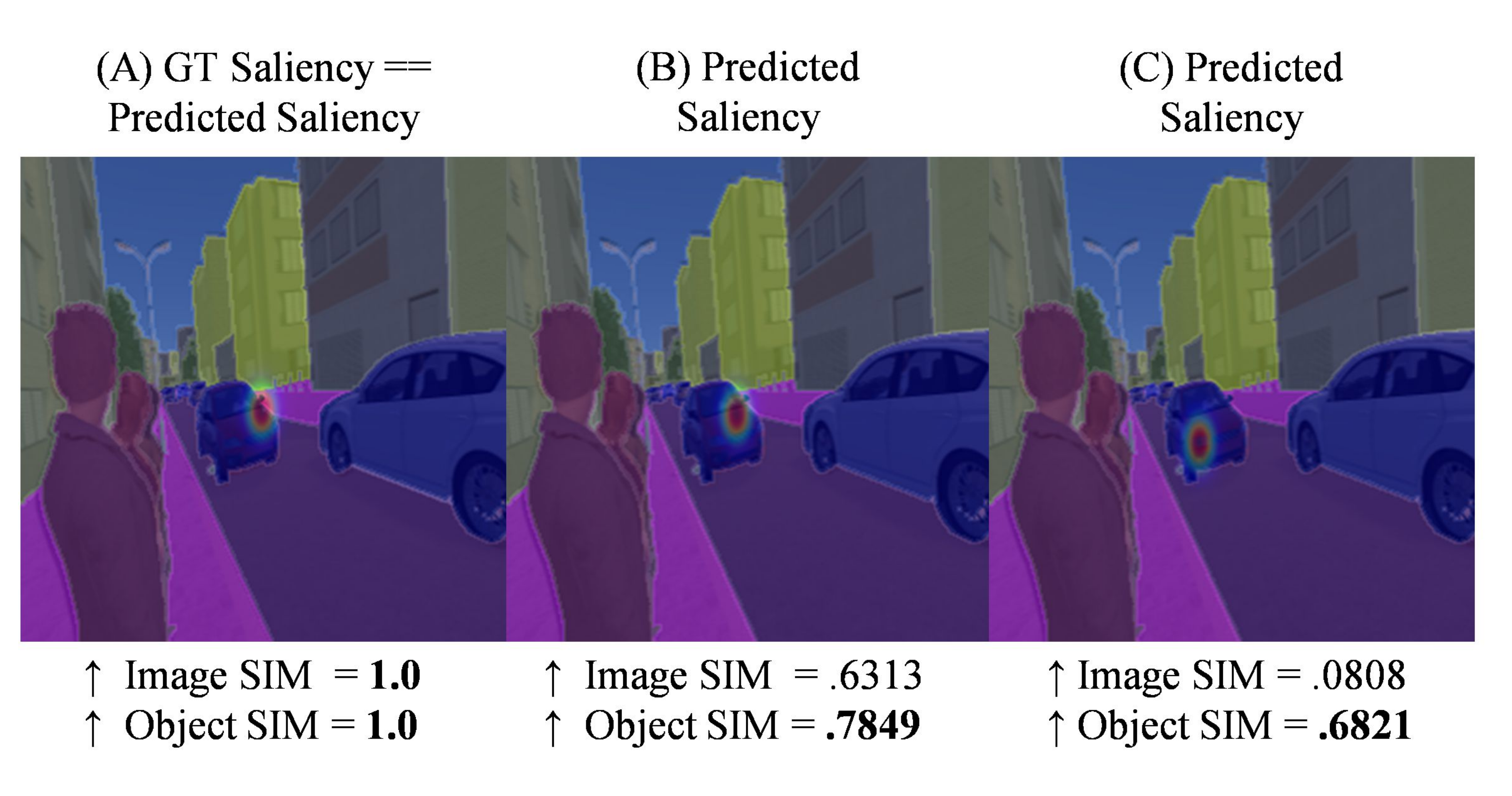}
   \caption{Comparison of oSIM with the SIM metric. Column (A) shows perfect alignment with the ground truth (SIM = oSIM = 1.0). In (B–C), predictions diverge, and both metrics decline. Unlike SIM, which drops quickly due to spatial misalignment, oSIM also accounts for object-level semantics (e.g., safety-critical approaching vehicle).
   }
   \label{fig:metric}
\end{figure}

We introduce Object-based Similarity (oSIM), the first metric to specifically evaluate object-based attention prediction, relevant for safety-critical settings like pedestrian street-crossing and its interaction with the approaching vehicle (cf. Figure~\ref{fig:metric}).
Object-based Similarity is based on the classical Similarity (SIM) metric used to evaluate saliency predictions given image space. Importantly, instead of operating on the raw pixel values without any notion of objects and its semantics, oSIM operates on the level of objects. In essence, the panoptic-based histograms are being compared based on the aggregated sum of pixel intensities across different classes:
\begin{equation}
    oSIM = \sum_{mask \in Image} \min \left( \sum_{i \in mask} S_i, \sum_{i \in mask} S^{gt}_i \right)
    \label{eq:oSIM_v2}
\end{equation}
where  \( mask \in Image \) is a mask corresponding to a single object in the image. $\sum_{i \in mask} S_i$ corresponds to the predicted saliency sum over pixels $i$ in the $mask$, where $\sum_{i \in mask} S^{gt}_i$ stands for the ground truth saliency sum over pixels $i$ in the $mask$, respectively.
Intuitively, instead of measuring similarity between predicted and ground truth attention on the image pixels, oSIM measures their similarity on the objects basis. 
As such, each object has the same influence on the overall oSIM metric, irrespective of its size.
In Figure~\ref{fig:metric}, we illustrate the differences between the classic image Similarity (SIM) and the object-based Similarity (oSIM) scores. As shown in columns (B-C), when the predicted saliency remains within the same object as the ground truth, oSIM yields a significantly higher value.
In this way, oSIM accounts for the relevance of objects for the human perceptual system (approaching vehicle while crossing in Figure~\ref{fig:metric}), whereas SIM is solely based on spatial image alignment. The proposed metric is applicable beyond the present context and can be extended to other research domains (e.g., driver attention, human–robot collaboration). Beyond panoptic-level segmentation, oSIM supports hierarchical, part-level segmentation, enabling objects to be decomposed into semantically meaningful internal components (e.g., body parts  of pedestrians).

\section{Methodology}
\label{sec:methodology}

\begin{figure*}[t]
  \centering
   \includegraphics[width=1\linewidth]{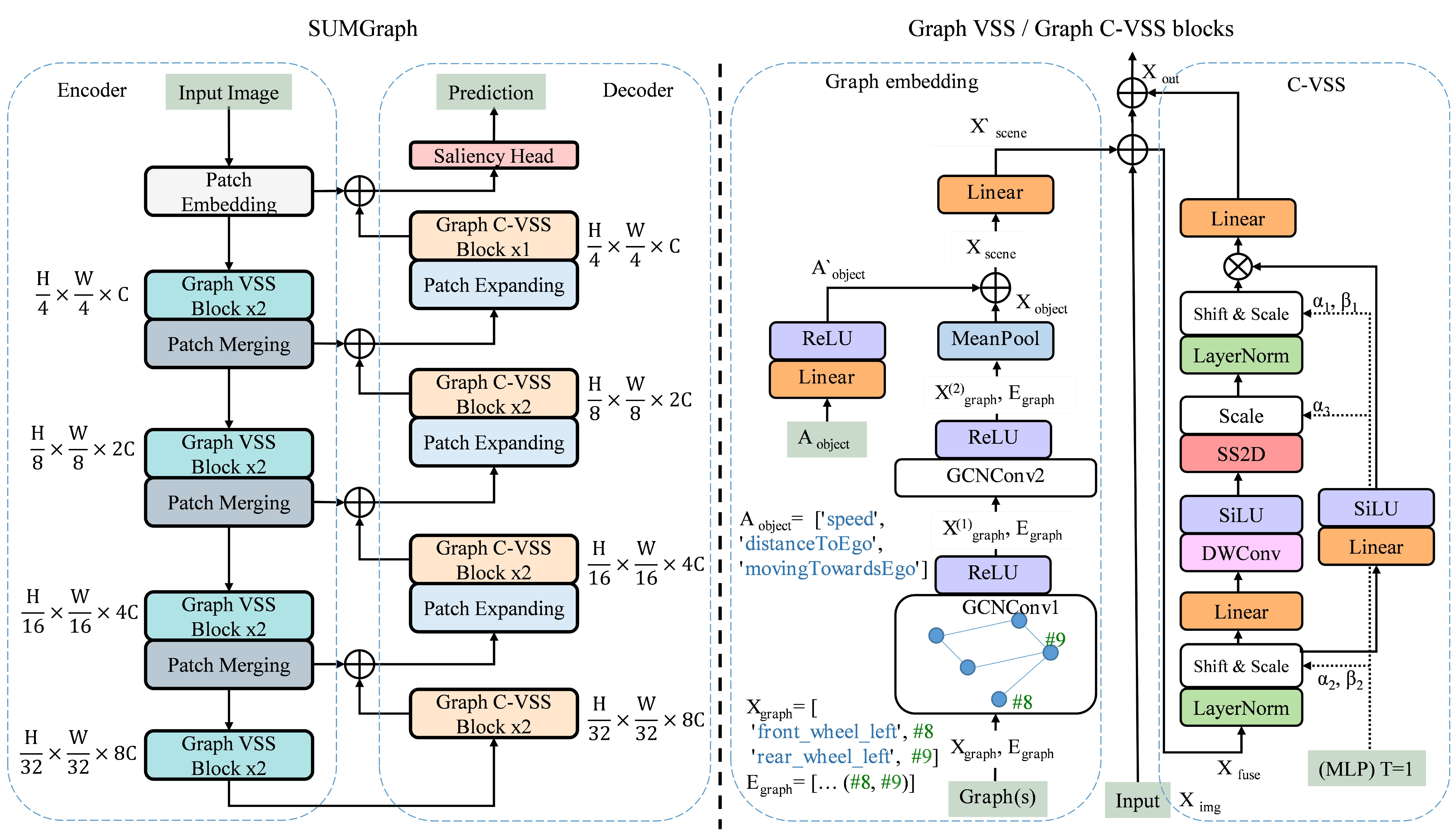}

   \caption{Left: Overall architecture of the proposed SUMGraph model for visual attention prediction. Right:  novel Graph VSS (fusion of graph and image features, in encoder) and Graph C-VSS (fusion of graph and conditional image features, in decoder) blocks for integration of additional contextual information. $\otimes$ stands for the element-wise produce operation, $\oplus$ is the element-wise addition. }
   \label{fig:CVSS_graph}
\end{figure*}

\textbf{Overview.} The overall architecture of SUMGraph is shown in Figure~\ref{fig:CVSS_graph} (left). Our approach builds upon the recent Mamba-U-Net-based model of~\cite{hosseini2025sum} and introduces novel Graph VSS and Graph C-VSS blocks, which extend the Visual State Space (VSS) and Conditional Visual State Space (C-VSS) blocks with a context graph that captures object-based scene information (cf. Figure~\ref{fig:CVSS_graph}, right). In the SUM architecture, the C-VSS block provides a unified scene representation by encoding semantic and spatial structure, while an attention mechanism selectively focuses on the most relevant objects and regions to support interaction-aware downstream reasoning, those characteristics were transferred to SUMGraph. The encoder produces four hierarchical output representations, where each block is followed by a downsampling layer that halves the spatial dimensions and doubles the number of channels. The decoder contains four Graph C-VSS blocks with two sub-blocks each, whereas the last block contains a single Graph C-VSS block. The patch-expanding layers perform upsampling to match the corresponding initial resolution, and the linear saliency head layer generates the output.

\textbf{Graph C-VSS block.} The novel graph-empowered Graph C-VSS block is shown in Figure~\ref{fig:CVSS_graph} (right). \( G~=~(V, E, A) \) denotes the graph structure for each object in the scene, where \( V \) represents nodes (e.g. vertices of vehicles), \( E \) represents edges (connections between vertices), \( A \) corresponds to the global object-based attributes, such as the speed, the distance to the user, and object's direction (towards the user or away).
\( X_{\text{graph}} \in \mathbb{R}^{M \times f} \) represents node features of the graph, where \( M \) is the number of vertices in each object's skeleton, and \( f \) is the feature dimension of each node.
\( A_{\text{object}} \in \mathbb{R}^{p} \) is the global attribute vector for each object, where \( p \) stands for the number of attributes.
\( E_{\text{graph}} \) denotes the edge index representing the graph structure for each object, where \( C \in \mathbb{R}^{c} \) is the residual condition vector for conditional modulation~\cite{hosseini2025sum}, with \( c \) dimensions.
In the first step (cf. Figure~\ref{fig:CVSS_graph}, Graph embedding), we process object skeletons and local attributes (e.g. \text{front\_wheel\_left}, as 
 in~\cite{kreiss2021openpifpaf}) using two consecutive graph convolutions. 
We apply the initial graph convolution to transform node features, followed by the second graph convolution to refine node embeddings:
\begin{equation}
\label{eq:GCNN}
\begin{aligned}
     X_{\text{graph}}^{(1)} = \sigma(\text{GCNConv}(X_{\text{graph}}, E_{\text{graph}}))
     \\ 
     X_{\text{graph}}^{(2)} = \sigma(\text{GCNConv}(X_{\text{graph}}^{(1)}, E_{\text{graph}}))
\end{aligned}
\end{equation}
where \( X_{\text{graph}}^{(1)}, X_{\text{graph}}^{(2)} \in \mathbb{R}^{M \times h} \) represent the hidden states with \( h \) as the number of hidden dimension, and \( \sigma \) stands for ReLU activation function. Later, we project the object's global attribute vector \( A_{\text{object}} \) to the same hidden dimension \( h \): $A_{\text{object}}^{\prime}~=~\sigma(W_{\text{attr}} A_{\text{object}})$, where
\( W_{\text{attr}} \in \mathbb{R}^{h \times p} \) is a learnable weight matrix. Upon completion, we aggregate node features within each object's skeleton using global mean pooling:
\begin{equation}
\label{eq:GCNN_2}
\begin{aligned}
     X_{\text{object}} = \text{MeanPool}(X_{\text{graph}}^{(2)}) + A_{\text{object}}^{\prime}
\end{aligned}
\end{equation}
where \( X_{\text{object}} \in \mathbb{R}^{h} \) is the combined feature vector for each object, integrating both node and global attribute information.
To obtain a scene-level context vector $X'_{\text{scene}}$, we aggregate and project across all object embeddings:
\begin{equation}
\label{eq:GCNN_3}
\begin{aligned}
     X_{\text{scene}} = \frac{1}{K} \sum_{k=1}^{K} X_{\text{object}}^{(k)}   \qquad    X_{\text{scene}}^{\prime} = W_{\text{scene}} X_{\text{scene}}
\end{aligned}
\end{equation}
where \( K \) stands for the number of objects. The scene-level context vector $X'_{\text{scene}}$ is fused with the visual features $X_{\text{img}}$ obtained from the C-VSS block as described in~\cite{hosseini2025sum}.

\textbf{Conditional VSS.}  
Lastly (cf. Figure~\ref{fig:CVSS_graph}, C-VSS), the conditional modulation step remains unchanged~\cite{hosseini2025sum}, allowing for adaptations to various domains. 
The C-VSS block is in line with~\cite{hosseini2025sum}, which allows for further extensions since the modulation of the graph-empowered feature maps through dynamic scaling and shifting operations are adjusting feature activations. $\alpha_i$ is a scaling factor, $\beta_i$ is a shifting factor, where the modulated feature map is equal to $\alpha_i \odot \text{(original feature map)} + \beta_i $. In this work, we used a single set token $T=1$. The token is fed into the Multi-Layer Perceptron (MLP) model, allowing it to be conditioned on the characteristics of natural scene-based eye tracking data type. The main objective of MLP is to regress $\alpha_i$ and $\beta_i$ parameters based on the data types~\cite{hosseini2025sum}. The regressed parameters dynamically adjust the model's behaviour as in~\cite{hosseini2025sum}, enhancing its performance and generalization.

\textbf{Loss function.}
Following previous work~\cite{hosseini2025sum}, our model employs a loss function that integrates several components~\cite{aydemir2023tempsal, droste2020unified, lou2022transalnet, hosseini2025sum}. 
We extend the loss with the $oSIM$ term to optimize for object-based attention explicitly: 
\begin{equation}
\label{eq:loss_final}
\begin{split}
    Loss = \lambda_1 \cdot {KLD}(S^{gt}, S) + \lambda_2 \cdot CC(S^{gt}, S) &\ + \\ \lambda_3 \cdot SIM(S^{gt}, S) + \lambda_4 \cdot NSS(F^{gt}, S) &\ +
    \\ \lambda_5 \cdot MSE(S^{gt}, S) + \boldsymbol{\lambda_6 \cdot oSIM(S^{gt}, S, P)}
\end{split}
\end{equation}
where $S^{gt}$ stands for the ground truth attention map, $S$ corresponds to the predicted attention map, $F$ denotes the binary fixation map, and $P$ is the panoptic segmentation. The $\lambda_i$ are the weighting coefficients. 
Each term targets a specific evaluation criterion commonly used to compare predictions of human attention maps to ground truth~~\cite{aydemir2023tempsal, droste2020unified, lou2022transalnet, hosseini2025sum}. 
$KLD$ stands for Kullback-Leibler Divergence,
$CC$ for the correlation coefficient, $SIM$ for the Similarity metric, $NSS$ for the Normalized Scanpath Similarity, and $MSE$ denotes Mean Squared Error.

\section{Experiments}
\label{sec:results}

\begin{table*}[ht]
\centering
\caption{The upper section presents results for models without fine-tuning on \dataset, while the middle section includes fine-tuned models, including ours. SUMGraph (with scale) refers to the explicit up/downsampling of graph information (see Figure~\ref{fig:CVSS_graph}, left). The third evaluation highlights performance gains on \dataset obtained by integrating the oSIM loss into the overall training objectives.}
\label{table:resultsMain}

\scriptsize
\begin{adjustbox}{width=\textwidth}
\begin{tabular}{l>{\centering\arraybackslash}p{1.1cm}>{\centering\arraybackslash}p{1.2cm}>{\centering\arraybackslash}p{.9cm}>{\centering\arraybackslash}p{.9cm}>{\centering\arraybackslash}p{.9cm}>{\centering\arraybackslash}p{.9cm}}
\toprule
Method & CC $\uparrow$ & KLD $\downarrow$ & AUC $\uparrow$ & SIM $\uparrow$ & NSS $\uparrow$ & oSIM $\uparrow$ \\
\midrule

\rowcolor{blue!20}\textit{1. w/o finetuning} \\
\ \ TempSAL~\cite{aydemir2023tempsal} & 0.0234 & 3.9136 & 0.6497 & 0.0412  & 0.1823 & 0.3457 \\
\ \ TranSalNet~\cite{lou2022transalnet} & 0.2340 & 2.9500 & 0.8812 & 0.1231 & 2.1626 & 0.3766 \\
\ \ ContextSalNet~\cite{Vozniak_2023_WACV} & 0.0093 & 3.9165 & 0.6177 & 0.0406  & 0.0760 & 0.3484 \\
\ \ SUM~\cite{hosseini2025sum} & 0.2961 & 2.6820 & 0.9160 & 0.2003 & 3.0513 & 0.4304 \\

\midrule

\rowcolor{blue!20}\textit{2. w/ finetuning} \\
\ \ TempSAL~\cite{aydemir2023tempsal} & 0.4342 & 1.8510 & 0.9606 & 0.3192 & 5.9976 & 0.5767 \\
\ \ TranSalNet~\cite{lou2022transalnet} & $0.4348^{\text{3rd}}$ & 1.7059 & 0.9672 & 0.3414 & 6.4809 & 0.5961 
\\
\ \ ContextSalNet~\cite{Vozniak_2023_WACV} w/~\cite{hosseini2025sum} loss & 0.4335 & 1.6941 & 0.9680 & 0.3462 & $\textbf{6.6447}$ & $0.6042^{\text{2nd}}$ \\
\ \ SUM~\cite{hosseini2025sum} & \textbf{0.4722} & 1.7062 & 0.9662 & 0.3470 & 6.2607 & 0.5909 \\

\rowcolor{gray!20} \textbf{SUMGraph (Ours)} & 0.4564 & $1.6747^{\text{2nd}}$ & \textbf{0.9683} & \textbf{0.3568} & 6.4357 & \textbf{0.6086} \\
\textbf{SUMGraph (Ours) (no global attr.) } & 0.4508 & 1.6729 & 0.9681 & 0.3482 & 6.3898 & 0.6019 \\
\rowcolor{gray!20} \textbf{SUMGraph (Ours) \& Scale} & $0.4643^{\text{2nd}}$ & \textbf{1.6581} & \textbf{0.9683} & $0.3481^{\text{2nd}}$ & $6.5241^{\text{2nd}}$ & 0.6025 \\

\midrule

\rowcolor{blue!20}\textit{3. w/ finetuning \& oSIM loss} \\
\ \ TempSAL~\cite{aydemir2023tempsal}  & \cellcolor{gray!20}{+0.0044} & +0.0006 & \cellcolor{gray!20}{+0.0101} & -0.0029 & \cellcolor{gray!20}{+0.0943} & \cellcolor{gray!20}{+0.0022} \\
\ \ TranSalNet~\cite{lou2022transalnet} & \cellcolor{gray!20}{+0.0135} & \cellcolor{gray!20}{-0.0234} & \cellcolor{gray!20}{+0.0002} & \cellcolor{gray!20}{+0.0085} & \cellcolor{gray!20}{+0.1648} & \cellcolor{gray!20}{+0.0072} \\
\ \ ContextSalNet~\cite{Vozniak_2023_WACV} w/~\cite{hosseini2025sum} loss \hspace{0.5cm}
& \cellcolor{gray!20}{+0.0044} & \cellcolor{gray!20}{-0.0098} & \cellcolor{gray!20}{+0.0005} & \cellcolor{gray!20}{+0.0011} & -0.0652 & \cellcolor{gray!20}{+0.0012} \\
\ \ SUM~\cite{hosseini2025sum}  & -0.0034 & \cellcolor{gray!20}{-0.0424} & \cellcolor{gray!20}{+0.0019} & \cellcolor{gray!20}{+0.0001} & \cellcolor{gray!20}{+0.1112} & \cellcolor{gray!20}{+0.0097} \\
\textbf{SUMGraph (Ours)} & -0.0023 & \cellcolor{gray!20}{-0.0130} & \cellcolor{gray!20}{+0.0001} & \cellcolor{gray!20}{+0.0062} & \cellcolor{gray!20}{+0.0148} & \cellcolor{gray!20}{+0.0045} \\

\bottomrule
\end{tabular}
\end{adjustbox}
\end{table*}

\textbf{Implementation details.}
To model the relevant scene aspects in street-crossing scenarios, we encode all vehicles that are in the participant's field of view into an object graph.
In particular, depending on their visibility to the user, we encode up to 24 2D keypoints as nodes in a graph.
Each node represents a local attribute, such as \textit{front\_wheel\_left}, following the labelling scheme presented in~\cite{kreiss2021openpifpaf}. Furthermore, each graph in the scene contains global attributes: speed, distance, and direction.
It is important to note that graph processing is performed only when there are relevant objects within the pedestrian's FoV. If no objects are present, the system defaults to the baseline model~\cite{hosseini2025sum}.

\textbf{Train/validation/test splits.} We divided the data into training, validation, and test sets using a 70\%, 10\%, and 20\% allocation strategy, which corresponds to 84, 12, and 24 participants, respectively. To avoid introducing biases in the training, we ensure an equal number of male/female and Japanese/German participants within the splits. The splits were verified for demographic balance using a Kolmogorov–Smirnov test on key variables (age and height).

\textbf{Training details.} We implemented our approach in the PyTorch framework and trained it on a cluster with 10 $\times$ A100 (80vGB) for 15 epochs with early stopping after 4 epochs. We used Adam optimisation, with the initial learning rate set to $1\times10^{-4}$, with a learning rate scheduler that decreased learning rate by a factor of 10 after four epochs. We employed Distributed Data-Parallelization (DDP), where the overall batch size is set to 750 samples. 
In line with ~\cite{hosseini2025sum}, we scale the resolution to 256$\times$256. 
The optimal values for the weighting coefficients \( \lambda_i \) in the loss function (cf. Equation~\ref{eq:loss_final}) are set to: \( \lambda_1 = 10 \), \( \lambda_2 = -2 \), \( \lambda_3 = -1 \), \( \lambda_4 = -1 \), \( \lambda_5 = 1 \), \( \lambda_6 = -1 \) and correspond to~\cite{hosseini2025sum}, whereas the $\lambda_6$ is determined through multiple test trials. The MLP architecture's layers match the baseline~\cite{hosseini2025sum}. For hidden layers within the graph embedding network, we apply Kaiming uniform initialization. This method is specific for layers utilizing ReLU activations, maintaining the variance of input signals and preventing issues like vanishing or exploding gradients~\cite{he2015delving}. 
Our SUMGraph architecture uses SUM~\cite{hosseini2025sum} weights pre-trained on six different datasets~\cite{jiang2015salicon, judd2009learning, borji2015cat2000, jiang2022does, jiang2023ueyes, xu2014predicting}. 
\begin{figure*}[pth!]
  \centering
   \includegraphics[width=1\linewidth]{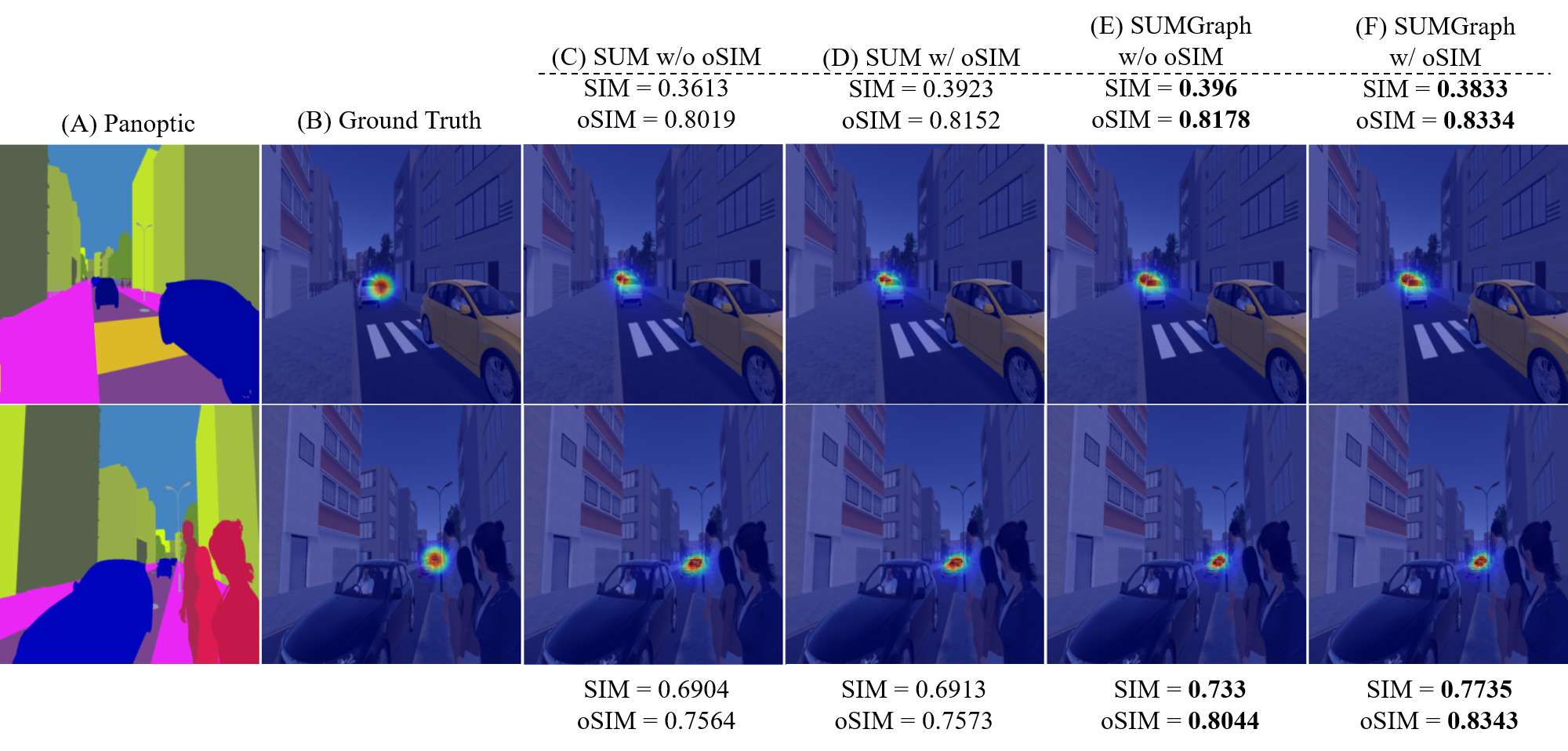}
   \caption{Qualitative visualizations of random samples. For clarity in performance improvements, both SIM and oSIM scores are provided. Column (A) presents the panoptic labels, (B) shows the ground truth attention map. The SUMGraph model (D) demonstrates improved performance compared to the baseline, with the incorporation of oSIM (columns C vs. D, and E vs. F) leading to consistent improvements.}
   \label{fig:qual}
\end{figure*}

\textbf{Metrics.} 
In line with previous work on visual saliency- and human attention prediction~\cite{Vozniak_2023_WACV, lou2022transalnet, hosseini2025sum, kummerer2016deepgaze, jiang2015salicon, wang2017deep}, we measure the quality of predicted attention with both location- and distribution-based metrics~\cite{bylinskii2018different}. Location-based metrics, such as $NSS$ and $AUC$ (Area Under the ROC Curve), represent ground truth with a binary fixation map. Distribution-based metrics, like $CC$, $SIM$, and $KLD$, evaluate the similarity between predicted and ground truth attention distributions. In addition, we evaluate predictions with novel $oSIM$ metric.

\subsection{Quantitative results}
We present comparisons against state-of-the-art saliency prediction approaches in Table~\ref{table:resultsMain}. We make use of the weights published by the original authors for initialization, except in the case of ContextSalNet~\cite{Vozniak_2023_WACV}, where these weights were not available and we needed to fall back to VGG~\cite{simonyan2014very} weights. The low performance achieved by models without fine-tuning highlight the difference between classical saliency estimation and prediction of sparse attention maps on \dataset. The top section of Table~\ref{table:resultsMain} presents the default performance of models without fine-tuning. In the middle section of Table~\ref{table:resultsMain}, we show the results of models after training on \dataset dataset.
SUMGraph performs on par with state-of-the-art models and achieves the best performance in the KLD, AUC, SIM, oSIM metrics. In summary, \textbf{SUMGraph outperforms prior methods in 28 out of 30 metrics}, demonstrating its effectiveness across both pixel- and object-level benchmarks. Moreover, the inclusion of the additional oSIM loss term as part of the total loss function contributes to a \textbf{performance boost in 25 out of 30 metrics} (cf. Table~\ref{table:resultsMain}, 3rd evaluations). This ablation study highlights not only the value of SUMGraph’s architectural design but also the general applicability, robustness, and consistency of our object-centric (oSIM) loss formulation. Furthermore, we investigate the impact of global attributes in our novel Graph (C-)VSS blocks. 
Our findings show that incorporating global attributes in the graphs improves the performance of SUMGraph. We also highlight the difference between the default and explicit graph structuring, where graphs are scaled to match block resolution.

\subsection{Qualitative results}
In Figure~\ref{fig:qual}, we randomly selected samples to showcase the performance of our method versus the baseline. As observed (cf. Figure~\ref{fig:qual}, columns C vs. D, E vs. E), the \textbf{utilization of oSIM as an additional training objective improves performance, while also resulted in improving SIM score.} To conclude, it guides the model's predictions toward capturing the semantic context of the object more accurately, which is especially relevant for safety-critical settings like street-crossing. \textbf{SUMGraph shows better performance in comparison to SUM}~\cite{hosseini2025sum}, reflecting the influence of the used graph structure (aligned with how pedestrians estimate approaching vehicles and its affiliated features) in safety-critical scenarios like street-crossing.

\section{Limitations and Future Work}
While our novel dataset, evaluation metric, and method improve the state-of-the-art in visual attention prediction, several limitations remain. 
Our training data is predominantly urban-centric, which may impact generalization to rural, extreme-weather environments or daylight conditions. 
Future work should mitigate these shortcomings.
The participants in our study were aged between 20 and 50 years, and the generalization to other age groups remains to be investigated.
While our dataset features participants from two distinct cultural backgrounds (DE and JP), a wider scope of backgrounds should be investigated.

\section{Conclusion}
\label{sec:conclusion}

In this work, we introduced the~\dataset~dataset, a novel VR resource applicable to diverse tasks, with a focus on visual attention prediction in street-crossing scenarios, an area challenging due to complexity, safety and ethical considerations. We proposed an object-based Similarity (oSIM) metric to capture object-driven attention, offering a perspective more aligned with human perception. Empirical results show that integrating oSIM into the loss function improves performance and advances modeling of visual attention. We also introduced the SUMGraph model, which exploits contextual graph information to improve predictive accuracy and achieve state-of-the-art performance.

\section{Acknowledgment}
This work was supported by the European Regional Development Fund (EFRE; EFRE-AuF-0000866) and in part by the European Union’s Horizon Europe research and innovation programme (No. 101076360). The initial study recording has been funded by the German Ministry for Research and Education (BMBF; 01IW17003), and partly by the New Energy and Industrial Technology Development (NEDO; JPNP18010). 
\bibliographystyle{IEEEtran}
\bibliography{biblatex}

@String(CVPR= {IEEE Conf. Comput. Vis. Pattern Recog.})

@String(ICCV= {Int. Conf. Comput. Vis.})

@String(TIP  = {IEEE Trans. Image Process.})

@String(TVCG  = {IEEE Trans. Vis. Comput. Graph.})

@String(ICLR = {Int. Conf. Learn. Represent.})

@String(VR   = {Vis. Res.})

@String(CVPR  = {CVPR})

@String(ICCV  = {ICCV})

@String(TIP   = {IEEE TIP})

@String(TVCG  = {IEEE TVCG})

@String(ICLR  = {ICLR})

@InProceedings{Cordts_2016_CVPR,
author = {Cordts, Marius and at al.},
title = {The Cityscapes Dataset for Semantic Urban Scene Understanding},
booktitle = {Proceedings of the IEEE CVPR},
month = {June},
year = {2016}
}

@article{chun2005visual,
  title={Visual attention},
  author={Chun, Marvin M and Wolfe, Jeremy M},
  journal={Blackwell handbook of sensation and perception},
  year={2005},
  publisher={Wiley Online Library}
}

@article{salvucci2001integrated,
  title={An integrated model of eye movements and visual encoding},
  author={Salvucci, Dario D},
  journal={Cognitive Systems Research},
  year={2001},
  publisher={Elsevier}
}

@inproceedings{sood2023improving,
  title={Improving neural saliency prediction with a cognitive model of human visual attention},
  author={Sood, Ekta and at al.},
  booktitle={CogSci},
  year={2023}
}

@article{fahimi2021metrics,
  title={On metrics for measuring scanpath similarity},
  author={Fahimi, Ramin and Bruce, Neil DB},
  journal={Behavior Research Methods},
  volume={53},
  pages={609--628},
  year={2021},
  publisher={Springer}
}

@article{recasens2015they,
  title={Where are they looking?},
  author={Recasens, Adria and Khosla, Aditya and Vondrick, Carl and Torralba, Antonio},
  journal={Advances in neural information processing systems},
  year={2015}
}

@inproceedings{chen2021predicting,
  title={Predicting human scanpaths in visual question answering},
  author={Chen, Xianyu and Jiang, Ming and Zhao, Qi},
  booktitle={Proceedings of the IEEE/CVF CVPR},
  year={2021}
}

@inproceedings{yang2020predicting,
  title={Predicting goal-directed human attention using inverse reinforcement learning},
  author={Yang, Zhibo and et al.},
  booktitle={CVPR},
  year={2020}
}

@inproceedings{muller2020anticipating,
  title={Anticipating averted gaze in dyadic interactions},
  author={M{\"u}ller, Philipp and Sood, Ekta and Bulling, Andreas},
  booktitle={ACM Symposium on Eye Tracking Research and Applications},
  year={2020}
}

@article{schneider1995vam,
  title={VAM: A neuro-cognitive model for visual attention control of segmentation, object recognition, and space-based motor action},
  author={Schneider, Werner X},
  journal={Visual Cognition},
  year={1995},
  publisher={Taylor \& Francis}
}

@article{chen2012object,
  title={Object-based attention: A tutorial review},
  author={Chen, Zhe},
  journal={Attention, Perception, \& Psychophysics},
  year={2012},
  publisher={Springer}
}

@article{kreiss2021openpifpaf,
  title={Openpifpaf: Composite fields for semantic keypoint detection and spatio-temporal association},
  author={Kreiss, Sven and Bertoni, Lorenzo and Alahi, Alexandre},
  journal={IEEE Transactions on Intelligent Transportation Systems},
  year={2021},
  publisher={IEEE}
}

@inproceedings{hosseini2025sum,
  title={Sum: Saliency unification through mamba for visual attention modeling},
  author={Hosseini, Alireza and Kazerouni, Amirhossein and Akhavan, Saeed and Brudno, Michael and Taati, Babak},
  booktitle={IEEE/CVF WACV},
  year={2025},
  organization={IEEE}
}

@inproceedings{zhou2014object,
  title={Object detectors emerge in deep scene cnns},
  author={Zhou, Bolei and Khosla, Aditya and Lapedriza, Agata and Oliva, Aude and Torralba, Antonio},
  booktitle={Proceedings of the ICLR},
}

@InProceedings{Vozniak_2023_WACV,
    author    = {Vozniak, Igor and et al.},
    title     = {Context-Empowered Visual Attention Prediction in Pedestrian Scenarios},
    booktitle = {Proceedings of the IEEE/CVF WACV},
    year      = {2023},
}

@inproceedings{vozniak2020infosalgail,
  title={InfoSalGAIL: Visual Attention-empowered Imitation Learning of Pedestrian Behavior in Critical Traffic Scenarios.},
  author={Vozniak, Igor and Klusch, Matthias and Antakli, Andr{\'e} and M{\"u}ller, Christian},
  booktitle={IJCCI},
  year={2020}
}

@article{borji2015cat2000,
  title={Cat2000: A large scale fixation dataset for boosting saliency research},
  author={Borji, Ali and Itti, Laurent},
  journal={arXiv preprint arXiv:1505.03581},
  year={2015}
}

@inproceedings{jiang2015salicon,
  title={Salicon: Saliency in context},
  author={Jiang, Ming and Huang, Shengsheng and Duan, Juanyong and Zhao, Qi},
  booktitle={Proceedings of the IEEE CVPR},
  year={2015}
}

@inproceedings{droste2020unified,
  title={Unified image and video saliency modeling},
  author={Droste, Richard and Jiao, Jianbo and Noble, J Alison},
  booktitle={Computer Vision--ECCV 2020},
  year={2020},
  organization={Springer}
}

@article{cornia2018predicting,
  title={Predicting human eye fixations via an lstm-based saliency attentive model},
  author={Cornia, Marcella and Baraldi, Lorenzo and Serra, Giuseppe and Cucchiara, Rita},
  journal={IEEE Transactions on Image Processing},
  year={2018},
  publisher={IEEE}
}

@article{liu2018deep,
  title={A deep spatial contextual long-term recurrent convolutional network for saliency detection},
  author={Liu, Nian and Han, Junwei},
  journal={IEEE TIP},
  year={2018},
  publisher={IEEE}
}

@inproceedings{djilali2024learning,
  title={Learning Saliency From Fixations},
  author={Djilali, Yasser Abdelaziz Dahou and McGuinness, Kevin and O’Connor, Noel},
  booktitle={Proceedings of the IEEE/CVF WACV},
  year={2024}
}

@article{han2022survey,
  title={A survey on vision transformer},
  author={Han, Kai and et al.},
  journal={IEEE transactions on pattern analysis and machine intelligence},
  year={2022},
  publisher={IEEE}
}

@article{lou2022transalnet,
  title={TranSalNet: Towards perceptually relevant visual saliency prediction},
  author={Lou, Jianxun and Lin, Hanhe and Marshall, David and Saupe, Dietmar and Liu, Hantao},
  journal={Neurocomputing},
  year={2022},
  publisher={Elsevier}
}

@article{hu21fixationnet,
	title={FixationNet: Forecasting eye fixations in task-oriented virtual environments},
	author={Hu, Zhiming and Bulling, Andreas and Li, Sheng and Wang, Guoping},
	journal={IEEE Transactions on Visualization and Computer Graphics},
	year={2021},
	publisher={IEEE}}

@article{bylinskii2018different,
  title={What do different evaluation metrics tell us about saliency models?},
  author={Bylinskii, Zoya and Judd, Tilke and Oliva, Aude and Torralba, Antonio and Durand, Fr{\'e}do},
  journal={IEEE transactions on pattern analysis and machine intelligence},
  year={2018},
  publisher={IEEE}
}

@article{bernal2023d,
  title={D-SAV360: A Dataset of Gaze Scanpaths on 360° Ambisonic Videos},
  author={Bernal-Berdun, Edurne and et al.},
  journal={IEEE TVCG},
  year={2023},
  publisher={IEEE}
}

@inproceedings{xu2024panonut360,
  title={Panonut360: A Head and Eye Tracking Dataset for Panoramic Video},
  author={Xu, Yutong and at al.},
  booktitle={ACM Multimedia Systems Conference},
  year={2024}
}

@inproceedings{david2023salient360,
  title={The salient360! toolbox: Processing, visualising and comparing gaze data in 3d},
  author={David, Erwan and at al.},
  booktitle={ETRA},
  year={2023}
}

@inproceedings{xu2018gaze,
  title={Gaze prediction in dynamic 360 immersive videos},
  author={Xu, Yanyu and et al.},
  booktitle={IEEE CVPR},
  year={2018}
}

@inproceedings{wang2022salientvr,
  title={SalientVR: Saliency-driven mobile 360-degree video streaming with gaze information},
  author={Wang, Shibo and et al.},
  booktitle={MobiCom},
  year={2022}
}

@article{krizhevsky2017imagenet,
  title={ImageNet classification with deep convolutional neural networks},
  author={Krizhevsky, Alex and Sutskever, Ilya and Hinton, Geoffrey E},
  journal={Communications of the ACM},
  publisher={AcM New York, NY, USA}
}

@article{simonyan2014very,
  title={Very deep convolutional networks for large-scale image recognition},
  author={Simonyan, Karen},
  journal={arXiv preprint arXiv:1409.1556},
  year={2014}
}

@inproceedings{kummerer2017understanding,
  title={Understanding low-and high-level contributions to fixation prediction},
  author={Kummerer, Matthias and Wallis, Thomas SA and Gatys, Leon A and Bethge, Matthias},
  booktitle={ICCV},
  year={2017}
}

@inproceedings{he2016deep,
  title={Deep residual learning for image recognition},
  author={He, Kaiming and Zhang, Xiangyu and Ren, Shaoqing and Sun, Jian},
  booktitle={Proceedings of the IEEE CVPR},
  year={2016}
}

@article{hosseini2024brand,
  title={Brand Visibility in Packaging: A Deep Learning Approach for Logo Detection, Saliency-Map Prediction, and Logo Placement Analysis},
  author={Hosseini, Alireza and at al.},
  journal={Discover Applied Sciences},
  year={2025}
}

@article{hu2021ehtask,
  title={Ehtask: Recognizing user tasks from eye and head movements in immersive virtual reality},
  author={Hu, Zhiming and Bulling, Andreas and Li, Sheng and Wang, Guoping},
  journal={IEEE Transactions on Visualization and Computer Graphics},
  year={2021},
  publisher={IEEE}
}

@article{banerjee2024hot3d,
  title={HOT3D: Hand and Object Tracking in 3D from Egocentric Multi-View Videos},
  author={Banerjee, Prithviraj and et al.},
  journal={CVPR},
  year={2025},
}

@misc{lv2024aria,
title={Aria Everyday Activities Dataset},
author={Zhaoyang Lv and et al.},
year={2024},
eprint={2402.13349},
archivePrefix={arXiv},
primaryClass={cs.CV}
}

@inproceedings{judd2009learning,
  title={Learning to predict where humans look},
  author={Judd, Tilke and Ehinger, Krista and Durand, Fr{\'e}do and Torralba, Antonio},
  booktitle={2009 IEEE 12th ICCV},
  year={2009},
  organization={IEEE}
}

@article{kummerer2016deepgaze,
  title={DeepGaze II: Reading fixations from deep features trained on object recognition},
  author={K{\"u}mmerer, Matthias and Wallis, Thomas SA and Bethge, Matthias},
  journal={arXiv preprint arXiv:1610.01563},
  year={2016}
}

@article{kratzer2020mogaze,
  title={MoGaze: A Dataset of Full-Body Motions that Includes Workspace Geometry and Eye-Gaze},
  author={Kratzer, Philipp and et al.},
  journal={IEEE RAL},
  year={2020}
}

@inproceedings{jiang2023ueyes,
  title={UEyes: Understanding visual saliency across user interface types},
  author={Jiang, Yue and at al.},
  booktitle={CHI},
  year={2023}
}

@inproceedings{jiang2022does,
  title={Does text attract attention on e-commerce images: A novel saliency prediction dataset and method},
  author={Jiang, Lai and et al.},
  booktitle={CVPR},
  year={2022}
}

@article{sitzmann2018saliency,
  title={Saliency in VR: How do people explore virtual environments?},
  author={Sitzmann, Vincent and et al.},
  journal={IEEE TVCG},
  year={2018},
  publisher={IEEE}
}

@article{celikcan2020deep,
  title={Deep into visual saliency for immersive VR environments rendered in real-time},
  author={Celikcan, Ufuk and Askin, Mehmet Bahadir and Albayrak, Dilara and Capin, Tolga K},
  journal={Computers \& Graphics},
  year={2020},
  publisher={Elsevier}
}

@inproceedings{he2015delving,
  title={Delving deep into rectifiers: Surpassing human-level performance on imagenet classification},
  author={He, Kaiming and Zhang, Xiangyu and Ren, Shaoqing and Sun, Jian},
  booktitle={Proceedings of the IEEE ICCV},
  year={2015}
}

@article{wang2023foveated,
  title={Foveated rendering: A state-of-the-art survey},
  author={Wang, Lili and Shi, Xuehuai and Liu, Yi},
  journal={Computational Visual Media},
  year={2023},
  publisher={TUP}
}

@article{vecera1994grouped,
  title={Grouped locations and object-based attention: Comment on Egly, Driver, and Rafal (1994).},
  author={Vecera, Shaun P},
  year={1994},
  publisher={American Psychological Association}
}

@article{roelfsema2006cortical,
  title={Cortical algorithms for perceptual grouping},
  author={Roelfsema, Pieter R},
  journal={Annu. Rev. Neurosci.},
  year={2006},
  publisher={Annual Reviews}
}

@inproceedings{aydemir2023tempsal,
  title={TempSAL-uncovering temporal information for deep saliency prediction},
  author={Aydemir, Bahar and Hoffstetter, Ludo and Zhang, Tong and Salzmann, Mathieu and S{\"u}sstrunk, Sabine},
  booktitle={Proceedings of the IEEE/CVF CVPR},
  year={2023}
}

@article{hu2020dgaze,
  title={Dgaze: Cnn-based gaze prediction in dynamic scenes},
  author={Hu, Zhiming and et al.},
  journal={IEEE TVCG},
  year={2020},
  publisher={IEEE}
}

@article{roth2023objects,
  title={Objects guide human gaze behavior in dynamic real-world scenes},
  author={Roth, Nicolas and Rolfs, Martin and Hellwich, Olaf and Obermayer, Klaus},
  journal={PLOS Computational Biology},
  year={2023}
}

@article{de2005semantic,
  title={Semantic effects on object selection in real-world scene perception},
  author={De Graef, Peter},
  year={2005},
  publisher={oxford University Press; Oxford}
}

@article{egly1994shifting,
  title={Shifting visual attention between objects and locations: evidence from normal and parietal lesion subjects.},
  author={Egly, Robert and Driver, Jon and Rafal, Robert D},
  journal={Journal of Experimental Psychology: General},
  year={1994},
  publisher={American Psychological Association}
}

@article{itti2002model,
  title={A model of saliency-based visual attention for rapid scene analysis},
  author={Itti, Laurent and Koch, Christof and Niebur, Ernst},
  journal={IEEE Transactions on pattern analysis and machine intelligence},
  year={2002},
  publisher={Ieee}
}

@article{wang2017deep,
  title={Deep visual attention prediction},
  author={Wang, Wenguan and Shen, Jianbing},
  journal={IEEE Transactions on Image Processing},
  year={2017},
  publisher={IEEE}
}

@inproceedings{kuemmerer2018salmetrics,
    author = {K\"{u}mmerer, Matthias and Wallis, Thomas S. A. and Bethge, Matthias},
    title = {Saliency Benchmarking Made Easy: Separating Models, Maps and Metrics},
    year = {2018},
}

@inproceedings{sprenger2023crosscdr,
  author={Sprenger, Janis and et al.},
  booktitle={IEEE IV}, 
  title={Cross-Cultural Behavior Analysis of Street-Crossing Pedestrians in Japan and Germany}, 
  year={2023},
  doi={10.1109/IV55152.2023.10186635}}

@article{xu2014predicting,
	author = {Xu, Juan and Jiang, Ming and Wang, Shuo and Kankanhalli, Mohan S. and Zhao, Qi}, 
	title = {Predicting human gaze beyond pixels},
	year = {2014}, 
	journal = {Journal of Vision},
	publisher={Association for Research in Vision and Ophthalmology}
}

\end{document}